# Manuscript Template

**Full Title**

A Novel Large Vision Foundation Model (LVFM)-based Approach for Generating High-Resolution Canopy Height Maps in Plantations for Precision Forestry Management

**Short Title**

Generating High-Resolution Canopy Height Maps based on a LVFM Approach

**Authors**

Shen Tan[1,2], Xin Zhang[3,*,†], Liangxiu Han[3,*,†], Huaguo Huang[1], Han Wang[4]

**Affiliations**

1. State Key Laboratory of Efficient Production of Forest Resources, Beijing Forestry University, Beijing 100083, China

2. Engineering Research Center of Carbon Sequestration of Forest and Grassland, Ministry of Education

3. Department of Computing and Mathematics, Manchester Metropolitan University, Manchester M1 5GD, UK

4. Department of Earth System Science, Ministry of Education Key Laboratory for Earth System Modelling, Institute for Global Change Studies, Tsinghua University, Beijing 100084, China

   *Address correspondence to: Xin Zhang (X.Zhang@mmu.ac.uk) and Liangxiu Han (L.han@mmu.ac.uk)

   †: These authors contributed equally to this work

**Abstract**

Plantations for producing profitable products play a critical role in supporting local livelihoods, which requires accurate monitoring to guide sufficient management activities. Additionally, recent expansion of plantations for sequestering atmospheric $CO_2$ motivated by China Certified Emission Reduction (CCER) guideline further underscore the urgent need for accurate and cost-effective methods to estimate plantation aboveground biomass (AGB). High-resolution canopy height maps (CHMs) are essential for capturing detailed plantation traits for AGB estimation, especially given the typically small scale of plantations. While airborne or unmanned aerial vehicle (UAV) based lidar remains the gold standard for acquiring high-resolution CHMs, its high-cost limits widespread use. With advancements in deep learning, predicting CHMs using remote sensing RGB data has emerged as a cost-effective alternative, though challenges remain in accurately extracting canopy height-related features. To address these challenges, we develop a novel model for high-resolution CHM generation based on Large Vision Foundation Model (LVFM). This model integrates a feature extractor, a self-supervised feature enhancement module to avoid spatial details loss in feature extraction, and a height estimator to produce high-resolution CHMs. Tested in the Fangshan District of Beijing, China—a region characterized by small, fragmented plantation parcels—our model, utilizing high-resolution (1-meter grid) RGB imagery from Google Earth, demonstrated superior performance compared to existing methods, including conventional convolutional neural networks (CNNs) and naive LVFM implementations. The model achieved a mean absolute error of 0.09 m, a root mean square error of 0.24 m, and a correlation coefficient of 0.78 when evaluated against lidar-based CHM observations in pixel-wise assessments. Our model also exhibits satisfactory performance when being generalized into non-training regions. Additionally, the CHMs generated by our model enabled over 90% success in individual tree detection and showed high accuracy in AGB estimation, and a reasonable performance in tracking plantations' growth. Our approach offers a promising tool for evaluating carbon sequestration in plantations and natural forests covering a large region.

# 1. Introduction

Plantations cultivate trees with agricultural precision, following standardized planting protocols and employing effective management techniques to optimize growth efficiency, making a significant contribution to the global economy. These plantations, which encompass a variety of categories such as fruit, oil, rubber, and timber production, form a critical sector that sustains the livelihoods of millions. Beyond their significant economic value, many plantations play a pivotal role in mitigating climate change by serving as key strategies for sequestering atmospheric $CO_2$, thereby contributing to efforts to combat global warming [1-3]. Notably, young plantations demonstrate greater carbon sequestration capacities compared to old-growth forests, significantly advancing current carbon reduction initiatives and climate change mitigation efforts [4-6]. Contemporary policy initiatives increasingly promote plantations as a means to achieve carbon targets by emphasizing their financial benefits [7, 8]. Simultaneously, plantations sometimes change natural forests with complex species combination into a monocultural forest, which increases the risk of carbon leakage yet decreases the biodiversity [9]. Rigorous assessment of tree growth and carbon sequestration in plantations is not only essential for traditional silviculture management but is also critical for carbon trading and for predicting forests' adaptability to climatic variations [3, 6, 10, 11].

The vertical structure of plantations presents a significant challenge for assessment using remote sensing (RS) imaging techniques. While optical sensors are effective for mapping regional forest and plantation distributions, they often fail to accurately capture structural traits and aboveground biomass (AGB) due to their limited capacity to penetrate the canopy [12, 13]. Consequently, the observed reflectance tends to correlate with AGB rather than directly indicate it, resulting in a diminished capacity to detect local AGB variability, particularly in dense forests where the saturation effect occurs [14]. Conversely, vegetation optical depth (VOD), a by-product of soil moisture retrieval, incorporates information on canopy water content and has been widely used for estimating forest AGB over large spatial scales [15-18].

As a derivative of RS observations, canopy height map (CHM) is a wall-to-wall image where each pixel represents the maximum canopy height, essentially providing a projection of the canopy surface [19]. Given that both canopy height and vertical structure are crucial for understanding forest functionality, high-resolution CHMs serve as an effective tool in various forestry applications, particularly in estimating forest AGB from continental-scale analyses to localized plantation monitoring [20-25].

Due to the typically small size of plantations, high-resolution CHMs are crucial for capturing detailed traits of plantations, especially for AGB estimation. In general, the high-resolution CHMs can be obtained based on lidar because of its ability to accurately measure the three-dimensional structure of forests and vegetation. Basically, lidar systems emit laser pulses and measure the time it takes for them to return after hitting an object. This allows for extremely precise measurements of distances and, therefore, canopy heights [26]. When applied to forested areas, lidar can accurately determine the height of the canopy and other structural characteristics of vegetation. Airborne lidar systems generate point clouds with high density, thus facilitates the depiction of 3-dimentional canopy structural detailly [26]. Retrieving CHMs from airborne lidar point clouds by interpolating the maximum height within each grid is a conventional method for simplifying structural information and reducing data complexity. Given the high accuracy of tree height measurements from lidar-derived CHMs, tree growth on the order of $10^{-2}$ meters can be detected [27]. Despite this exceptional accuracy, generating CHMs over large areas with airborne lidar typically requires professional instruments, making it nearly impossible to collect lidar-derived CHMs for historical periods, which limits the ability to track forest growth over time in specific regions. On the other hand, spaceborne lidar instruments detects the terrestrial canopy height by series of laser samples with larger footprints than airborne lidar, but covers a continental to global range [28]. A wall-to-wall CHM can be then extrapolated by fitting the relationship between canopy heights and optical reflectance [29].

Monocular depth estimation is a technique that aims to predict the depth of each pixel in a single RGB image. This method has gained attention and importance in computer vision due to it its cost and accessibility. Recent advances in deep learning and computer vision have significantly improved the accuracy of monocular depth estimation. The model such as Convolutional Neural Networks (CNNs) and Vision Transformer (ViT) can learn to infer depth from various visual cues in an image, such as texture, shading, and perspective. These methods have also been used to predict CHMs using RGB remote sensing data. For example, training a CNN-based network with approximately 600 million Global Ecosystem Dynamics Investigation (GEDI) samples enables the generation of global-scale CHMs with a spatial resolution of 10 meters [21]. Furthermore, CNN-based methods can produce CHMs with even finer resolutions. Li et al. [30] demonstrated this by training a CNN with around 34,500 hectares of lidar-derived CHM references to generate nationwide CHMs with a 10-meter grid from RGB imagery. To fully leverage lidar data from multiple platforms, a hybrid network trained with approximately 160,000 airborne lidar patch samples and informed by GEDI data facilitates the generation of 10-meter grid CHMs across a state-level area [31].

However, as outlined above, training deep learning models often demands a vast number of training samples, particularly a significant, yet sometimes unattainable, quantity of CHM references. Large Vision Foundation Models (LVFMs) such as Distillation Network for Object Recognition V2 (DINOv2) and Contrastive Language-Image Pretraining (CLIP) represent a recent breakthrough in computer vision and deep learning, characterized by self-supervised pre-training on diverse imagery datasets [32, 33]. LVFMs excel in tasks and datasets they weren't explicitly trained on, offering high adaptability and reducing the computational resources and time needed to develop specialized models [32]. Among the most prominent LVFM architectures (e.g., DINOv2 and CLIP) is the ViT, which adapts the transformer architecture from natural language processing to image processing [34-36]. ViTs partition input images into patches, treating them as tokens processed through attention layers, enabling the capture of long-range dependencies and global context across the image. These approaches have yield superior feature representations compared to CNNs, particularly for depth prediction tasks, and have significantly advanced feature extraction tasks from RS imagery, such as object detection, segmentation, and classification [37-40].

Moreover, ViTs have proven to be highly effective as feature extractors for CHM generation in dense-canopy regions [41, 42].

Nevertheless, a common drawback of ViTs as feature extractors is their tendency to reduce the spatial resolution of input images. By dividing images into patches and applying multiple self-attention layers, ViTs often lose fine-grained details and spatial accuracy, which can degrade the performance of downstream tasks such as AGB estimation and tree segmentation, particularly in small plantations. Overcoming this limitation is essential for effectively applying ViTs in generating accurate CHMs. Additionally, whether the globally trained ViT-based network performs acceptably in plantation applications that requires high-level accuracy remains unclear [43].

China's recent policies encourage afforestation through plantations by emphasizing the additional profits from carbon trading [8]. While lidar-based methods are recommended for AGB quantification, generating high-resolution CHMs from RGB imagery offers a more cost-effective alternative for continuous monitoring of tree growth, as RGB imagery can be collected by diverse space-borne platforms at a significantly recued cost. However, conventional LVFM-based CHM generation methods tend to lose feature details, leading to reduced accuracy when applied to small plantations. To address this issue, we develop a novel LVFM-based CHM generation network and evaluated its performance in this study. Our model is designed to extract generalizable features from high-resolution RGB imagery and limited training, producing CHMs specifically tailored for plantation areas. We introduce a multi-stage fusion decoder that integrates outputs from different feature stages to predict pixel-wise canopy height with high precision. The two main contributions of the work are as followers. Firstly, we develop a novel high-resolution CHM prediction model that consists of three key components: a feature extraction from a LVFM, a self-supervised feature enhancement technique to preserve spatial detail during and a lightweighted tree height estimator and demonstrate the effectiveness of the LVFM on RGB imagery, and demonstrate a state-of-the-art resulting accuracy in CHM generation compared to existing methods and products. Secondly, this study showcases the model's proficiency in estimating above-ground biomass for plantations, achieving satisfactory accuracy in individual tree segmentation and demonstrating strong correlations in biomass estimation and highlights the model's potential for continuous monitoring of tree growth in plantations, providing a scalable and cost-effective solution for long-term forestry and carbon sink monitoring.

2. **Materials and Methods**

In this study, a novel high-resolution CHM prediction model is developed, which integrates three main parts: Feature extraction from LVFM, Feature enhancement and Tree height estimator, enabling the generation of CHMs from high-resolution RGB imagery. We evaluate the model's performance and explore its applicability in monitoring plantation through three experiments (Fig. 1). Experiment I involve a systematic evaluation of our model, focusing on the resulting metrics and extracted spatial features. These outcomes are concurrently compared with various network configurations, other CHM generation methods, and conventional CHM products. In Experiment II, we test the applicability of the inferred CHM in AGB estimation, which includes two parts: tree segmentation for each plantation and AGB estimation based on tree heights. We also apply the model to a historical dataset to assess its practicality in tracking plantation growth over time. Detailed descriptions of the model structure, the data used, and the experimental design are provided in Sections 2.1, 2.2, and 2.3, respectively.

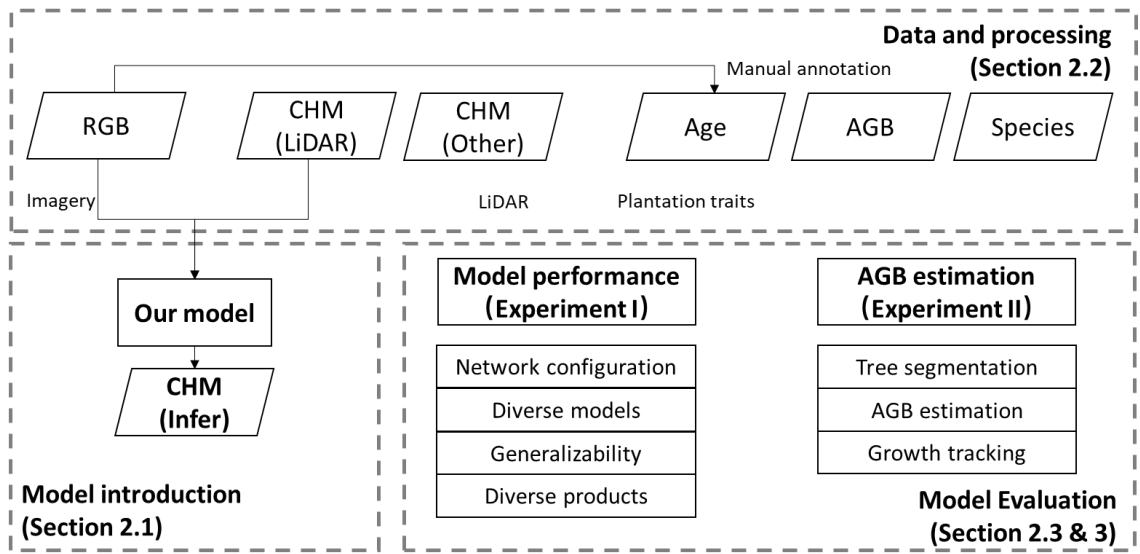

Fig. 1. Roadmap of this study. Correspond chapters are labelled in each part.

## 2.1 Method

In this work, we develop an innovative end-to-end canopy height estimation network. The system architecture overview is shown in Fig. 2. There are three main parts in the network: feature extractor with LVFM, the self-supervised feature enhancement module and tree height estimator. The rationale behind this network are as follows:

**Feature Extractor with LVFM:** CHM estimation from RGB images requires understanding the spatial arrangement and distance of vegetation within an image. The LVFM are pre-trained on vast amounts of diverse data, which helps them capture a wide range of spatial relationships and patterns. This extensive training allows them to generalize well to new, unseen data. Feature extracted from LVFMs can effectively learn the spatial patterns and contextual information. By leveraging the pre-trained knowledge from the LVFMs, CHMs estimation algorithms can achieve better performance, especially in generalizing across different environments and visual conditions.

**Self-supervised feature enhancement module:** ViTs which largely used in LVFMs, rely on self-attention mechanisms that operate on relatively large patches of the image, which can result in a loss of spatial details critical for CHMs prediction task. Therefore, we introduce a self-supervised feature enhancement module. This novel component enhances the resolution of extracted features through a self-supervised upscaling approach, which leverages low-resolution signals from multiple views to learn high-resolution representations.

**Tree Height Estimator:** A lightweight CNN module is designed to predict tree heights from the enhanced features. This estimator refines the high-resolution features, reducing their dimensionality and projecting them into the final CHM. The design is focused on computational efficiency while ensuring precise height predictions, making the entire network both effective and scalable.

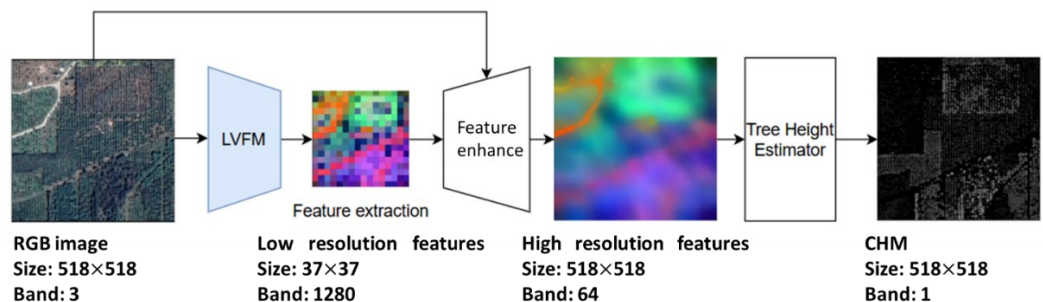

Fig. 2. Network structure in this study. This model consists of three main components: a LVFM as feature extractor, a feature enhancement module, and a tree height estimator. The size of low-resolution features (37×37) refers to the outputs of DINOv2.

### 2.1.1 Feature extractor

The feature extractor based on LVFMs adapts the vision transformer network, which divides the image into patches and treats them as tokens that are fed into a series of attention layers. This approach captures long-range dependencies and global context across the image, resulting in superior feature representations [44].

In this study, we tested several LVFMs as feature extractor, including: DINOV2, Masked autoencoder[45], CLIP [33], and Intel-DPT [34]. These models offer a broad spectrum of capabilities and methods for visual understanding. DINOV2 and Masked autoencoder represent advanced development in self-supervised learning for vision tasks. They are known for its ability to learn robust features from unlabelled data, making it highly versatile and efficient for various downstream tasks. CLIP has emerged as a powerful model that bridges vision and language. By jointly training on images and their textual descriptions, CLIP learns multimodal representations that are highly effective for tasks involving both visual and textual information, such as zero-shot learning, image captioning, and visual search. Intel-DPT is a LVFM that specifically designed for dense prediction tasks, such as depth estimation, segmentation, and object detection. According to Fig. 3, features extracted from all LVFMs exhibit some degree of mosaic distortion, as ViTs split the data into patches, resulting in a loss of spatial information. Among the LVFM-based feature extractors, DINOv2 employs a patch size of 14 for its image split, allowing it to capture finer details and provide higher resolution than other methods using a patch size of 16 [42]. Given the superior performance of DINOv2 across different network complexities and LVFM selections, we choose DINOv2 as our feature extractor in the following parts of this study.

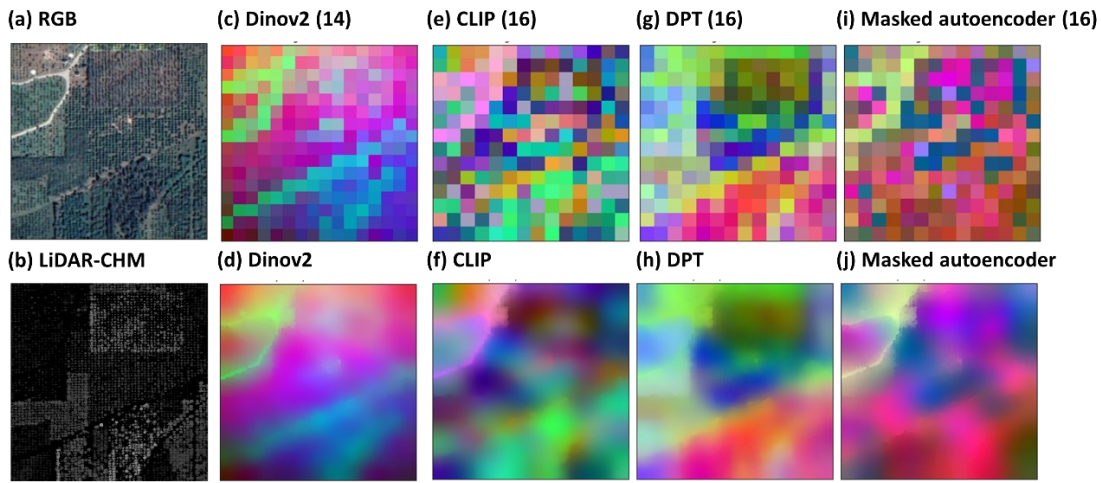

Fig. 3. Comparison between (a) RGB image, (b) CHM reference, and feature bands from four LVFMs. Panels (c, e, g, i) display the feature maps extracted from DINOv2, CLIP, DPT, and Masked autoencoder, respectively, panels (d, f, h, j) display the feature maps after feature enhancement. Patch size of each model is labelled in the panels respectively. Principal Component Analysis (PCA) is used to project the original multichannel data onto the three principal components for visualization.

### 2.1.2 Feature Enhancement

The feature enhancement module is a self-supervised feature upscaling method that enhances the feature map from the feature extractor without requiring external labeled data (Fig. 4). The process begins by generating multiple low-resolution features from different views. To achieve this, various augmentations (e.g., flipping, padding, cropping) are applied to the input images ($\{X_i \mid i = 1,2,...,n\}$, where $n$ represents the total number of patches). Low-resolution features ($LF_i$) are then extracted from the feature extractor, allowing the observation of subtle differences in the output features and providing sub-feature information to train the Feature Enh module, i.e., $LF_i = f(t(X_i))$.

Next, the feature enhancement module produces high-resolution features ($HF_i$) aligned with object edges. We adapt a lightweight method known as CARAFE (Content-Aware ReAssembly of FEatures) to upscale the feature in our model [46]. Then the same augmentation strategies are applied to these high-resolution features to generate high-resolution features across multiple views. A downsampler ($\sigma_\downarrow$) is used to transform $HF_i$ into low-resolution features: $\sigma_\downarrow(HF_i) = LF_i'$. The downsampler does not alter the "space" or "semantics" of the features with nontrivial transformations but rather interpolates features within a small neighborhood. The downsampler blurs features using a learned kernel, implemented as a convolution applied independently to each channel. The learned kernel is normalized to a non-negative sum of 1, ensuring that the features remain in the same space. We then compare the downsampled features to the original LF using a multiview consistency loss.

$$\mathcal{L}_{rec} = \frac{1}{|n|}\sum_n \frac{1}{2s^2} \|LF_i - LF_i'\|_2^2 + \log(s) \qquad (1)$$

where $\|.\|$ is the standard squared $l_2$ norm and $s = \mathcal{N}(LF_i)$ is a spatially varying adaptive uncertainty parameterized by a small linear network $\mathcal{N}$. This turns the mean-square error (MSE) loss into a proper likelihood capable of handling uncertainty[47]. This extra flexibility allows the network to learn when certain outlier features fundamentally cannot be up sampled.

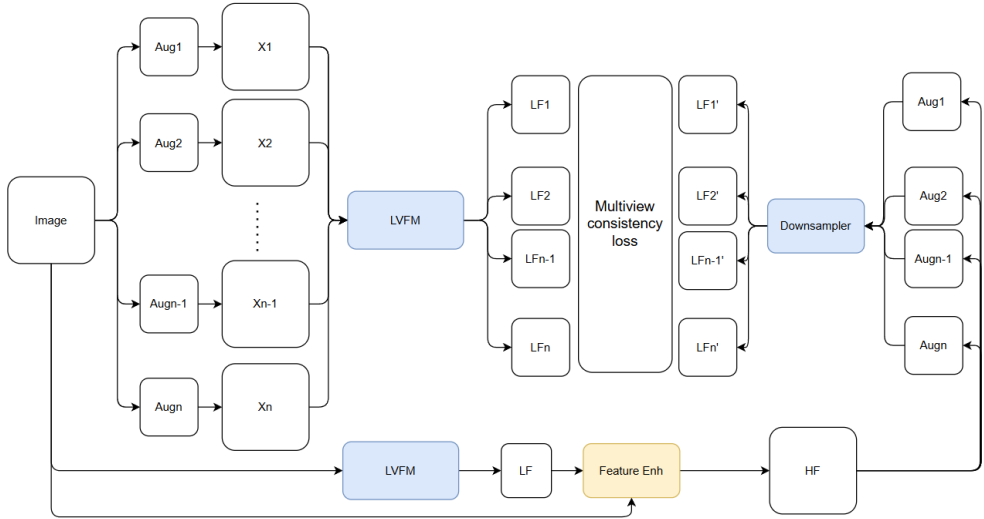

Fig. 4. Structure of the Feature enhancement module.

### 2.1.3 Tree height estimator

We use a CNN-based head to project tree height from the enhanced feature bands. This module consists of three convolutional layers, each followed by a batch normalization layer and an activation function. The design ensures that the model is lightweight, making it computationally efficient and suitable for deployment in environments with limited resources and labels (e.g., lidar references in this study). The architecture of this CNN head is specifically designed to efficiently reduce the dimensionality of features and transform them into the desired output range. Initially, two CNN layers with a kernel size of 3, followed by a ReLU (Rectified Linear Unit) activation function, are applied. Subsequently, a CNN layer with a kernel size of 1 is employed to generate the height estimates. Finally, a

Sigmoid function maps the output features into a range between 0 and 1. The process can be formulated as:

$$Height = Sig\text{Conv}_1(\text{Conv}_3(\text{Conv}_3(F_{hr}))) \qquad (2)$$

## 2.2 Material
### 2.2.1 Study area

The experiments were conducted in the Shilou County, Fangshan District of Beijing. A field campaign aiming at collecting lidar observation were finished in the northwest part of the study area (three lidar plots in the Fig. 5). In this region, a significant number of plantations have been established under the China Certified Emission Reduction (CCER) standard[7, 10]. The majority of these plantations were established in the early 2010s, with most cultivated either for the sale of saplings or for continuous $CO_2$ removal.

### 2.2.2 Data

Four types of data are used in this study: 1) time series sub-meter scale satellite RGB imagery as input, 2) lidar observations for model training as reference, 3) plantation traits, and 4) existing CHM products for comparison.

The satellite RGB images were sourced from Google Earth using the Bigemap tool (http://www.bigemap.com/), covering the period from 2013 to 2020 [48]. We have collected level-17 imagery with a spatial resolution of around 1.2 meters. Except for approximately 20% of the area in 2015, which was affected by cloud contamination, and 30% of the region in 2018, where imagery was only available during the winter season, all other RGB imagery was collected during the growing season (May to September).

The airborne lidar data are collected during a field investigation in November 2020 using a long-range laser scanner (RIEGL VUX-1LR) mounted on an unmanned aerial vehicle (UAV) system. The following equipment configurations were adopted during the investigation: flying height = 150 m, flying speed = ~6 m/s, lidar sampling frequency = 380 kHz, peak point cloud density = 100 points/m², beam divergence = 0.35 mrad, and ranging accuracy = 25 mm. The original lidar waveforms were decomposed into point clouds, and noise points were subsequently removed using a frequency histogram. The canopy and ground points were classified to create a digital terrain model (DTM) and a digital surface model (DSM) with a grid size of 1 m. The CHM was then calculated by subtracting the DTM from the DSM. Approximately 2.5 km² of CHM was generated (Fig. 5). For a more detailed description of the lidar data collection, refer to Qin et al. [10].

We have conducted another field investigation covering the entire study area in April 2024. This investigation aimed to collect data on tree species, guided by 1,436 manually annotated plantation parcels, including 212 parcels located within the three plots with lidar reference (indicated by red dots in Fig. 2a representing their centers). This data serves as the basis for evaluating the performance of our model inference. The majority of the plantations consist of economically beneficial species: 389 parcels of *Pinus tabulaeformis*, 192 of *Populus tomentosa*, 362 of *Robinia pseudoacacia*, 218 of *Ginkgo biloba*, 87 of *Salix matsudana*, 85 of *Fraxinus chinensis*, and 11 of *Toona sinensis*. The remaining 92 parcels contain species used for producing peaches or apricots. Additionally, the stand ages of each plantation are manually labeled based on their first appearance in historical RGB imagery. As a reference, we have also collected terrestrial greenness represented by the normalized difference vegetation index (NDVI) from Sentinel-2 imagery. The maximum NDVI value for each pixel from June to September 2020 was calculated, with cloud pixels detected and masked according to the method suggested by Frantz et al. [49].

Three existing CHM products are used in the evaluation experiment. One of these is a recent LVFM-based model for generating 1-meter CHMs by Tolan et al., which provides a global-scale CHM dataset [42]. This dataset was generated using multiple high-resolution imagery from 2018 to 2020 with less than 1-meter resolution, constrained globally by GEDI samples, making it comparable to the field investigation in this study. We also employ two widely used CHM products with 10- or 30-meters resolution from Potapov et al. and Lang et al. for comparison [21, 29]. The Potapov product was generated based on the non-linear relationship between optical signals, specifically from the Landsat series, and tree heights. The tree heights in the Lang product were estimated using a CNN-based network that considers the contribution of local pixels. Notably, both products were trained using CHM data from airborne lidar and GEDI observations.

To evaluate the model's generalizability, we further applied our trained network to a managed conifer forest in Saihanba (116°51'-117°39' E, 42°02'-42°36' N). This region is characterized by a cold temperate continental monsoon climate, marked by distinct seasonal variations, including harsh winters and dry, windy springs. The area exhibits an annual mean temperature of -1.2°C and receives approximately 500 mm of annual precipitation. Airborne lidar data, serving as the reference, was acquired in September 2018. The raw point cloud data underwent preprocessing steps, including noise removal and classification, before being interpolated into a 1-meter resolution CHM covering an area of approximately 5 km². High-resolution RGB imagery from Google Earth was acquired in August 2018.

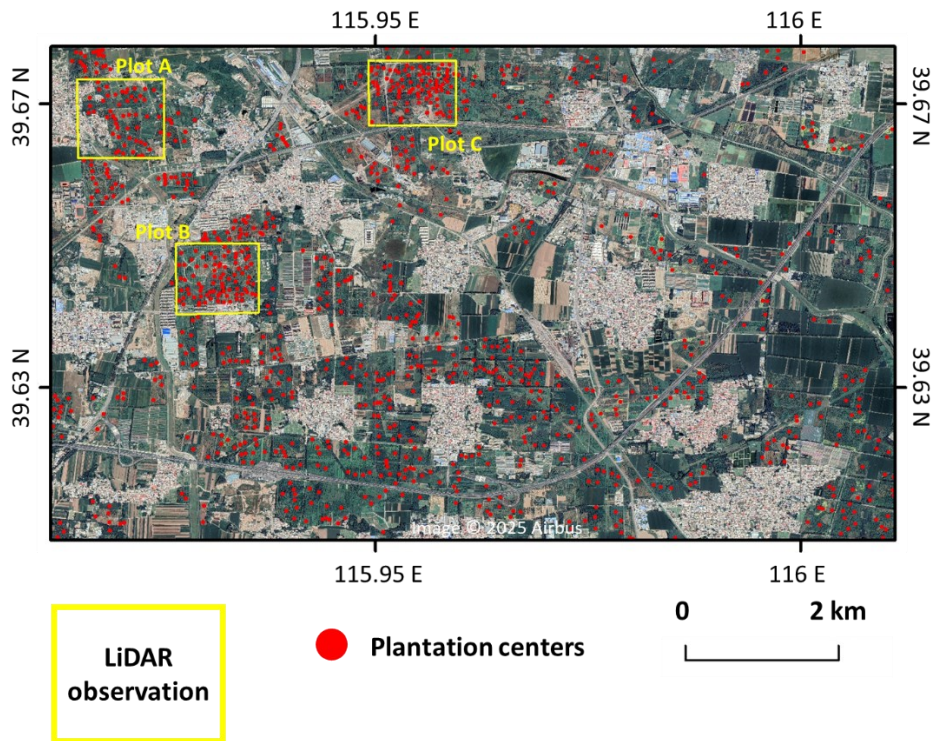

Fig. 5. RGB imagery of the study area. The position of three plots with lidar reference is marked by yellow rectangle. The centre of 1,436 plantation with species and age labels are represented by red dots.

## 2.3  Experiment design
### 2.3.1  Model configuration

The models used in this study are implemented using PyTorch and executed on an NVIDIA RTX A6000 GPU and an Intel(R) Xeon(R) W-2255 CPU. The training process was managed using the fastai framework [50], which was employed to determine the optimal learning rate. To avoid potential spatial autocorrelation, data from Plot A (2027×1753) and Plot C (1743×1659) were used for training, while data from Plot B (1813×1672) were used for evaluation. All tiles for training and evaluating the model were subset from the three plots without overlapping.

### 2.3.2  Experiments

We conduct two experiments in this study including: 1) Experiment I: Model performance evaluation; 2) Experiment II: Assessment of model suitability for the AGB estimation and tracking plantation growth.

1) **Experiment I—Model performance evaluation**

In this experiment, we firstly evaluate the model performance via various network configurations of the LFVM encoder, DINOv2, which offers several configurations of network complexity. We test three typical configurations: DINOv2-Small, DINOv2-Base, and DINOv2-Large. This comparison aims at selecting the best model configuration for the downstream tasks and analysis. A pixel-wise evaluation of the model outputs is performed by comparing the predicted CHMs against lidar observations using standard statistical metrics: mean absolute error (MAE), root mean squared error (RMSE), and correlation coefficients ($R^2$). Then, we compare the CHMs generated by two typical deep learning models: a CNN-based network, U-Net, and DPT, which has been partially adapted for CHM generation [42]. The comparison with U-Net aims at to prove the overperformance to typical CNN-based CHM generation method, while the comparison with DPT is designed to demonstrate the performance of our feature enhancement module. For this experiment, the network configuration for DPT is as follows: network type is Inter-DPT-Large, patch size is 16×16, embedding dimension is 1024, number of layers is 24, number of attention heads is 16, and the multilayer perceptron (feedforward dimension) is 4096. The configuration for U-Net is: network type is ResNet-34, with both encoder depth and decoder depth set to 5. Finally, we compare our CHM results with three typical CHM products introduced in Section 2.2.

During the evaluation of the model's generalizability, we applied the trained network to the Saihanba forest. For model inference, only a limited subset of lidar observations (approximately 5% of the total tiles) was utilized to project the extracted features into continuous tree height estimates. The resulting CHM was evaluated against lidar observation.

**Experiment II—Assessment of model suitability for the AGB estimation** Experiment II was also referenced against lidar observations within the three plots. We assess the applicability of using the model-generated CHM for AGB estimation. First, the number of trees in each plantation parcel was detected using a local maximum method with a radius of 5 meters [51]. We then estimate the average AGB within each plantation based on a modified function suggested by the CCER methodology (China Certified Emission Reduction (CCER), 2022). Since diameter at breast height (DBH) is typically required for AGB estimation but is not observable through routine RS methods, we use the relationship between tree height (H, in meters) and DBH (in centimeters) that has been fitted for the current study area (DBH = 1.117H + 5.38, see Qin et al. [10]). Non-linear AGB functions driven by tree heights are developed for each species (Table 1), allowing for the calculation of AGB for each plantation.

We also infer a CHM covering the entire study area and evaluated the potential of our model to track tree growth represented by the AGB increase from 2013 to 2020 for the main tree species.

Table 1. AGB estimation functions for main tree species in this study.

| Species | Functions |
| --- | --- |
| *P. Tabulaeformis* | AGB = $0.92H^2$-0.46H+5.03 |
| *P. Tomentosa* | AGB = $0.54H^2$-0.27H+2.97 |
| Other broadleaf species | AGB = $5.37H^2$-20.86H+33.95 |

3. **Results**
3.1 **Model performance evaluation**

### 3.1.1 Comparison between typical LVFMs and network configurations

A more complex network generates a more reliable CHM result (Table 2). DINOv2-Small, with its simpler network structure (embedding size = 384, depth = 12, number of attention heads = 6, and 21 million trainable parameters), produces less satisfactory accuracy: MAE = 0.14 meters, RMSE = 0.36 meters, and $R^2$ = 0.60. Increasing network complexity significantly enhances accuracy, as a larger embedding dimension allows for more image patches to be processed simultaneously, while greater network depth, more attention heads, and additional trainable parameters enrich the features that can be captured. Consequently, DINOv2-Large achieves the best inference performance, with an MAE of 0.09, RMSE of 0.24, and $R^2$ of 0.78. Additionally, the features extracted by more complex networks retain more spatial detail: the edges of tree crowns on the feature maps can only be clearly distinguished by DINOv2-Large (Fig. 6), forming a crucial basis for height prediction and tree detection in subsequent steps.

Table 2. Performance comparison on three typical ViT configurations of Dinov2: small, base, and large. The configurations in the header: Dim (embedding dimension), Depth (number of layers), Heads (number of attention heads), and Para (total number of trainable parameters, in million). Units of MAS and RMSE are all in meter(s).

|  | Configuration parameters | | | | Resulting metrics | | | |
| --- | --- | --- | --- | --- | --- | --- | --- | --- |
|  | Dim | Depth | Heads | Para | Bias | MAE | RMSE | $R^2$ |
| Small | 384 | 12 | 6 | 21 | -0.04 | 0.14 | 0.36 | 0.60 |
| Base | 768 | 12 | 12 | 86 | -0.04 | 0.13 | 0.26 | 0.65 |
| Large | 1024 | 24 | 16 | 300 | -0.03 | 0.09 | 0.24 | 0.78 |

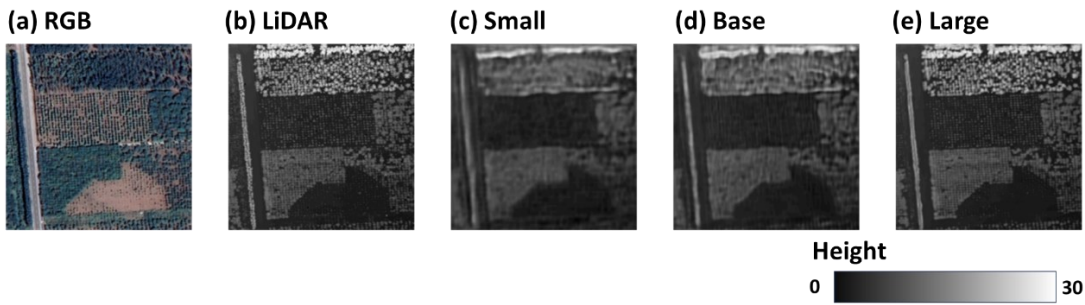

Fig. 6. Comparison between (a) RGB map, (b)lidar reference, and (c-e) CHM from three network configuration: DINOv2-Small, DINOv2-Base, and DINOv2-Large.

### 3.1.2 Model performance comparison with competitor models.

Overall, our model achieves the best performance in predicting tree heights among all three deep learning models, with the highest $R^2$ and the lowest prediction uncertainty (MAE and RMSE, as shown in Fig. 7 and Table 3). The U-Net model achieves a comparable $R^2$ of 0.76, but with greater uncertainty (MAE of 0.14 meters and RMSE of 0.33 meters). In contrast, the DPT-Large model exhibits the poorest performance among the three models. Given that the training scheme for all models was identical (80% of the lidar observations used as input and the remaining 20% as reference), the network structure is the key determinant of the resulting accuracy.

The U-Net architecture consists of a series of CNN kernels and pooling operations, whereas the ViT-based architectures within DINOv2 and DPT rely on patch-based tokenization and self-attention mechanisms. While these mechanisms prioritize global context and semantic features, they tend to reduce the emphasis on fine-grained local information, leading to the loss of some spatial details. This is evident in the DPT results, where nearly no tree crowns are distinguishable, suggesting that the tree heights predicted by DPT are closer to a local average rather than individual tree heights. This lack of distinction is particularly pronounced in plantations with significant tree height variability (see the middle section of Fig. 8a). On the other hand, CNN-based models like U-Net have smaller receptive fields than ViTs, causing the trained networks to focus more on local representations of the relationship between RGB signals and tree heights. However, this can lead to an overemphasis on local variability and a lack of consistency across entire plantations, resulting in the meaningless variability seen in Fig. 8b.

In addition to the overperformance of ViT-based network, we improved our model by incorporating the feature enhancement module, which avoid the spatial feature loss on the extracted features. The resulting maps shown in Fig. 8 illustrate the superiority of our approach: tree crowns are barely distinguishable in the U-Net and DPT results, whereas our model effectively preserves these details. Notably, these better-preserved crown details contribute to the significant improvement of the MAE and RMSE in Tables. Yet, as the edge part of crowns only takes limited part of the total area, there is only limited increase of resulting $R^2$.

Table 3. CHM prediction performance of three deep learning models. Units of MAE and RMSE are all in meter(s).

| Model | Bias | MAE | RMSE | $R^2$ |
|---|---|---|---|---|
| U-Net | -0.06 | 0.14 | 0.33 | 0.76 |
| DPT-Large | -0.03 | 0.14 | 0.36 | 0.71 |
| This study | -0.03 | 0.09 | 0.24 | 0.78 |

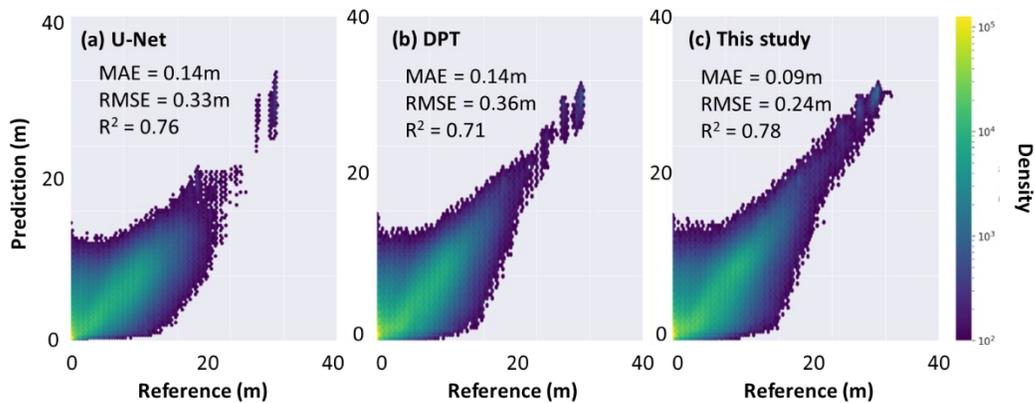

Fig. 7. Pixel-wise CHM evaluation between (a)U-Net, (b)DPT, and (c) proposed model in this study. Corresponding statistics are labelled in each panel respectively.

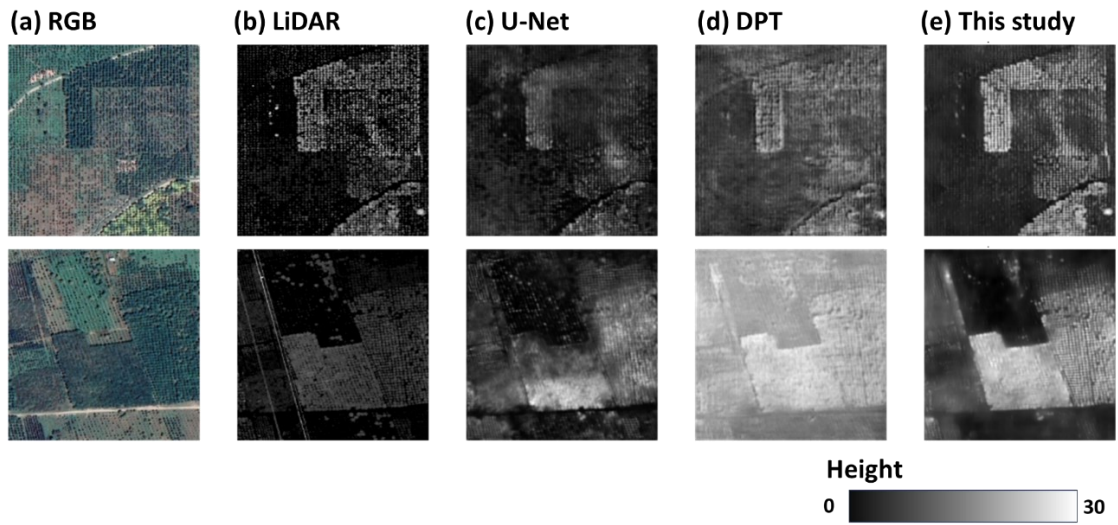

Fig. 8. Comparison of (a) RGB image, (b) LiDAR reference, and three CHM prediction from deep learning models: (c) U-Net, (d) DPT, and (e) model from this study.

### 3.1.3 Model generalizability

Quantitative analysis (Bias = -0.04, MAE = 0.13 meters, RMSE = 0.30 meters, and $R^2$ = 0.71) and visual inspection (Fig. 9) collectively indicate that the model exhibits satisfactory generalizability when applied to non-training regions. Despite the absence of intensive training in this area and the distinct forest structural traits compared to those in the Fangshan region, the trained model effectively discriminates between forests and shrublands, identifies individual old-growth tall trees, and distinguishes between mature and young forest stands.

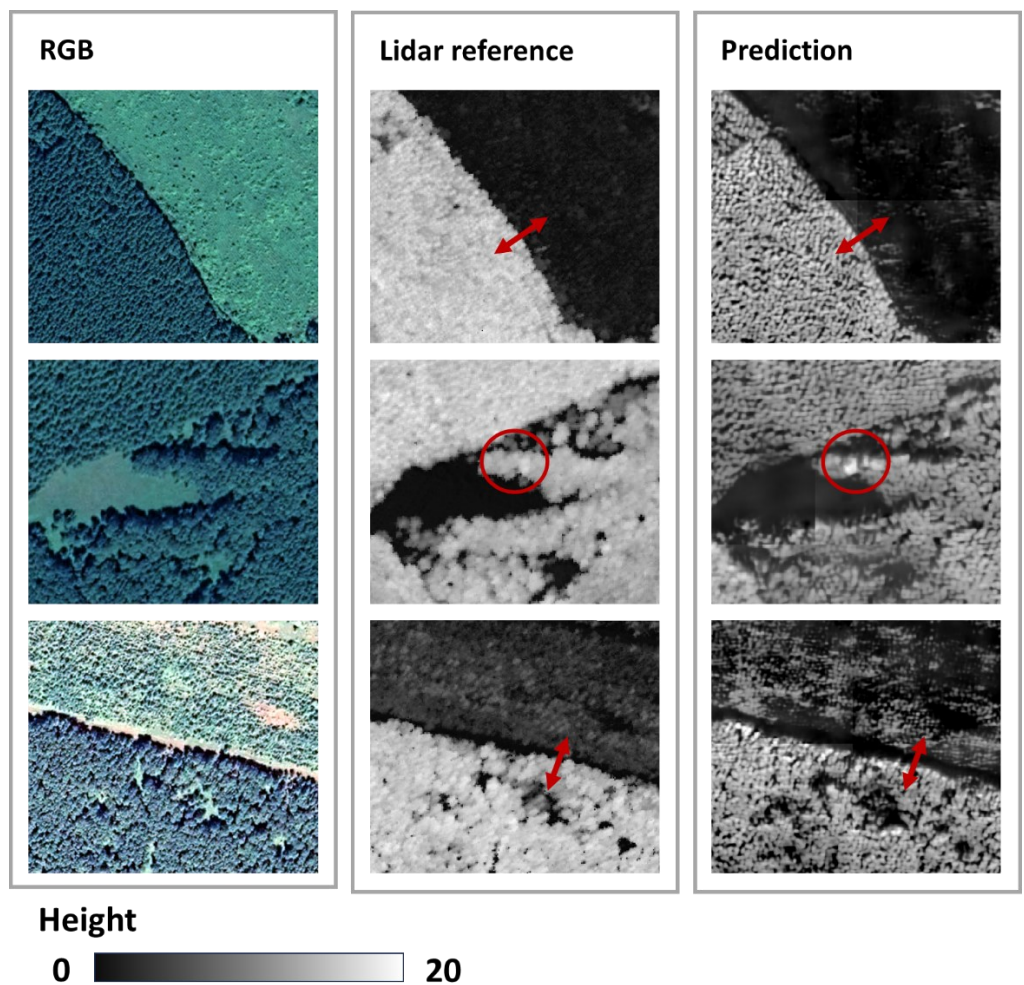

Fig. 9. Comparison between lidar reference and model prediction in three typical forest scenarios in Saihanba

### 3.1.4 CHM prediction performance comparison with typical products

Compared with another high-resolution CHM from Tolan et al. and two medium-resolution CHMs from Potapov et al. and Lang et al., our model demonstrates the closest similarity to the lidar observations (Fig. 10) [21, 29, 42]. Our results effectively capture the high tree height plantations in all three plots, particularly the shielding forests (mainly composed of Populus tomentosa) in the central part of Plot 3. In contrast, the 1m resolution CHM from Tolan et al. does not reflect strong variabilities in these regions, as the LVFM underlying this product loses part of the spatial features[42]. This suggests that Tolan's global-trained product may be less effective for monitoring AGB in plantations, which has also been suggested by recent study. Additionally, the two medium-resolution products poorly represent global tree height or distribution patterns in the resulting CHMs, failing to provide accurate tree height information. The zoomed-in maps in Fig. 11 further confirm that our model produces the best spatial patterns, while the other products do not capture the variability of tree heights within each plantation.

The CHM generated by our method shows the strongest correlation with the reference X or Y axis profiles in all three plots (Fig. 12). While Tolan's product, which utilizes features from high-resolution imagery, performs slightly better than the medium-resolution CHMs in most cases, it still falls short compared to our model. It is also noteworthy that the two medium-resolution CHMs exhibit a stronger correlation with the NDVI profile rather than with tree heights. Since NDVI represents greenness rather than structural or height information, it often fails to reflect local tree height variability accurately, which reduces the applicability of these optical-index-based CHM products (usually extrapolated from GEDI observations) in high standard plantation monitoring.

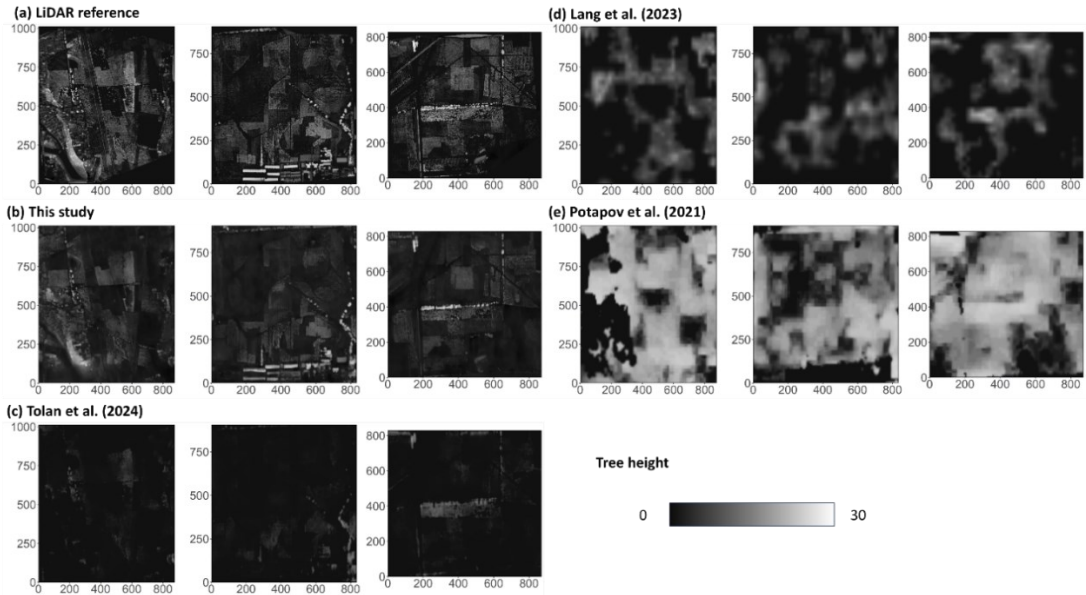

Fig. 10. Comparison of CHM result from (a) lidar observation, (b) this study, (c) Tolan et al., (d) Lang et al., and (e) Potapov et al. for the three plots with lidar reference[21, 29, 42]. X and Y axis represents the a projected coordinate system that set the lower left corner as the origin (0, 0).

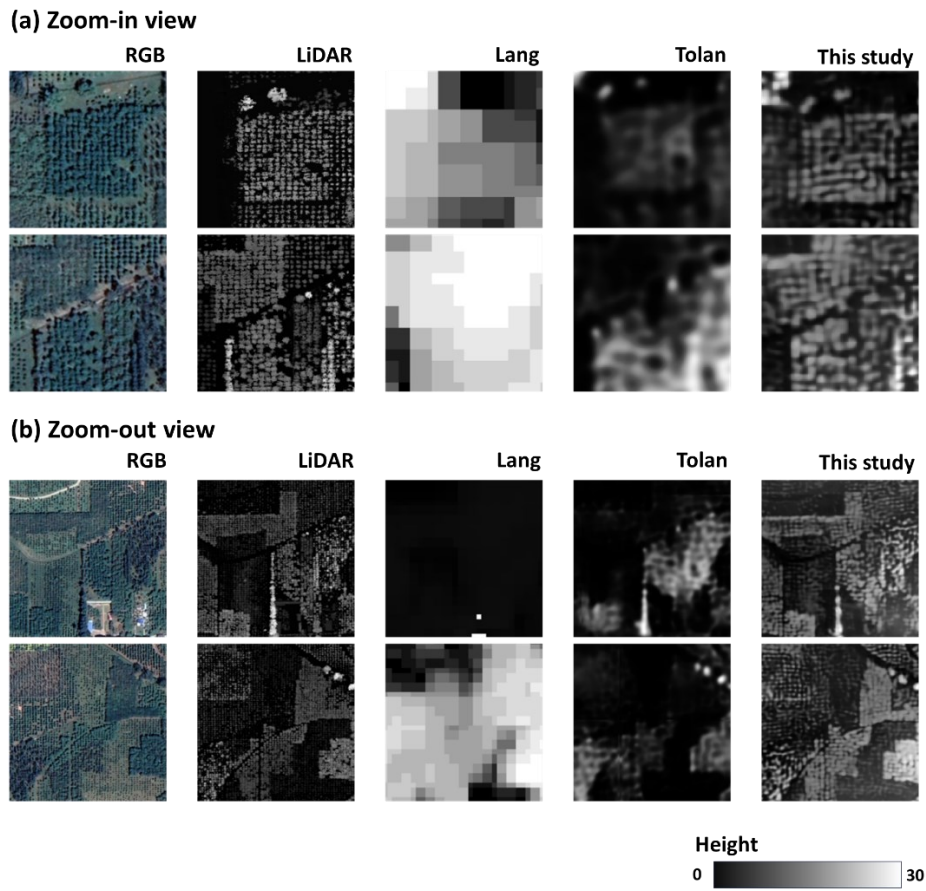

Fig. 11. Comparison of CHM from Lang et al. (third column) and Tolan et al. (fourth column) from this study (fifth column), with two zoom-in levels (a and b). The RGB map and lidar reference are displayed in the first and second column respectively[21, 42].

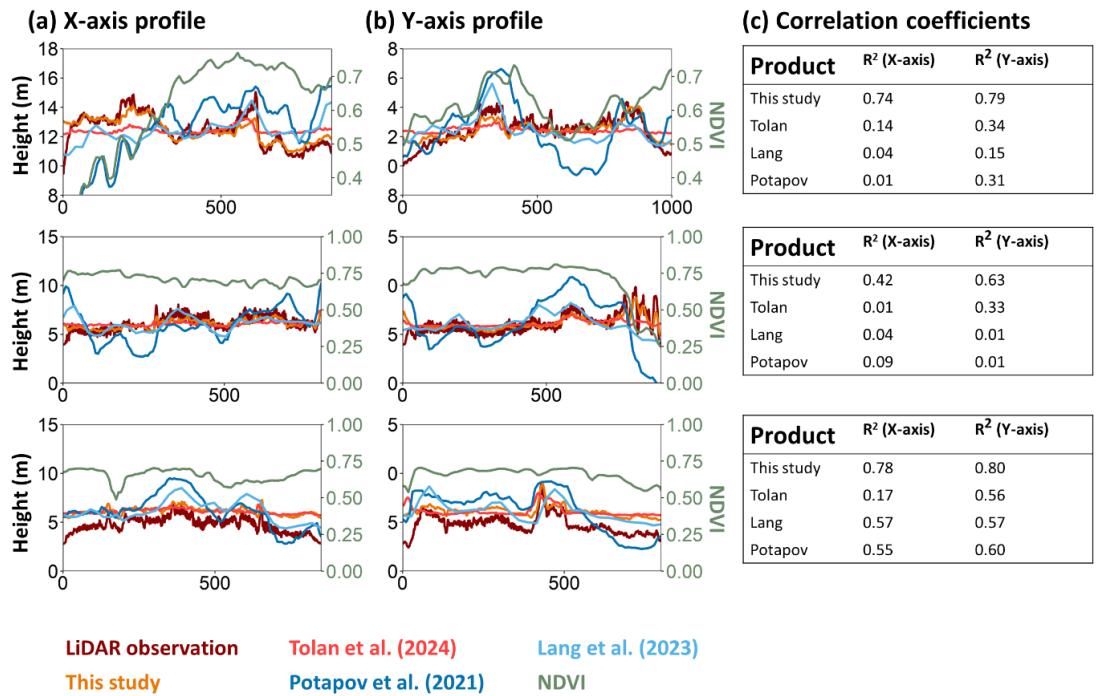

Fig. 12. (a) X, (b) Y profile, and (c) corresponding statistics for all CHM products within three plots. Profiles for different products are drawn by different colours.

## 3.2 Plantation AGB estimation

### 3.2.1 Tree segmentation based on CHM

Accurate tree detection, a prerequisite for estimating plantation AGB, can be reliably achieved using the CHM results from our model (Fig. S1). In two example regions, most trees are successfully detected, with success rates of 93% and 94%, respectively. Approximately 6% of the total trees, primarily those with limited height or closely grouped with adjacent trees, are not detected. This satisfactory performance demonstrates that our model effectively preserves fine-grained spatial details. For all 212 plantations with lidar reference data, tree detection based on our results maintains high accuracy, with an $R^2$ of 0.99 and a success rate of 92±6% (Fig. 13). The majority of plantations (135 out of 212) exhibit a success rate greater than 90%, with only a minimal rate of missed small trees, thereby enabling accurate AGB estimation at the plantation scale.

### 3.2.2 AGB estimation

Based on the CHM generation results, we estimate the average AGB for different plantations covering four main tree species (Fig. 14a). The $R^2$ values for these species indicate a strong correlation: 0.90 for *Pinus tabulaeformis*, 0.95 for *Populus tomentosa*, 0.71 for *Robinia pseudoacacia*, and 0.83 for *Ginkgo biloba*. These results demonstrate the reliability of using model-generated CHMs for estimating total AGB in each plantation (Fig. 14b).

Overall, the accuracy of plantation AGB estimation aligns with the performance of average AGB estimation. For plantations that are well-managed, such as those with *Pinus tabulaeformis*, *Populus tomentosa*, and *Ginkgo biloba*, the AGB estimation shows high reliability, with $R^2$ values of 0.94, 0.98, and 0.84, respectively. However, the AGB estimation for *Robinia pseudoacacia* shows reduced accuracy ($R^2 = 0.71$), likely due to some of these plantations being abandoned and no longer managed, leading to irregular features. Additionally, as a major plantation species in current study area, the AGB for Robinia pseudoacacia across different age groups are consistent with the lidar reference data (Fig. S2). As expected, an increase in plantation age generally corresponds to a continuous increase in AGB storage.

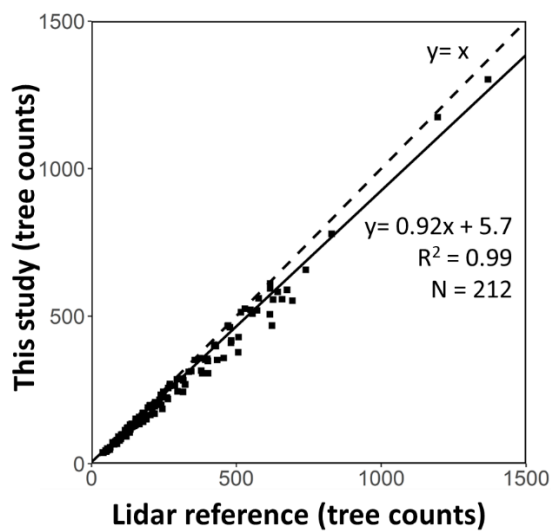

Fig. 13. Individual tree detection performance by CHM from lidar reference and this study. N represents the total number of plantations.

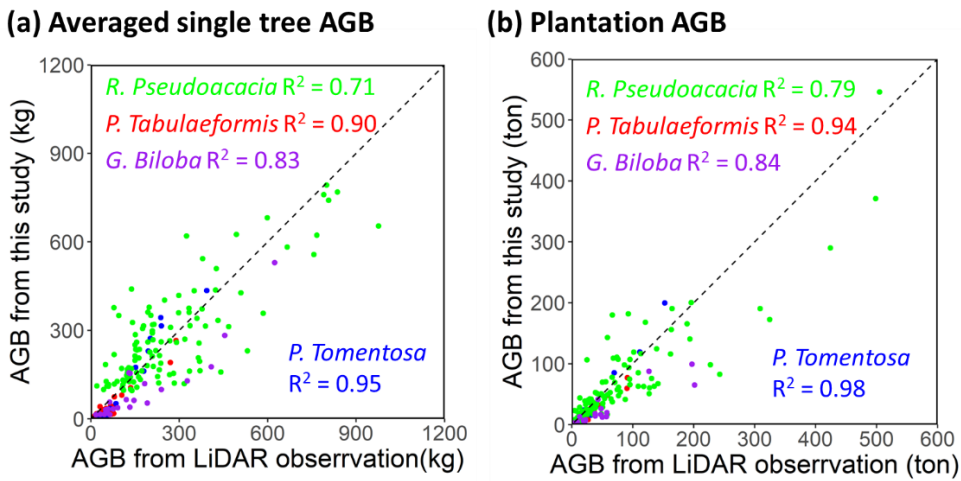

Fig. 14. (a) Averaged single tree and (b) plantation AGB estimation for each plantation based on CHM from lidar reference and this study. Tree species are represented by different colors.

### 3.2.3 Tracking plantation growth

We then estimate the tree growth rate for seven major species across the entire study area (Fig. S3). The afforestation project began around 2013, so a limited AGB increase during the early stages was expected. Over time, the AGB for all species show a significant increasing trend. Among these, *Salix matsudana* and *Toona sinensis* exhibited the highest growth efficiencies, with rates of 13.06 kg/yr and 15.18 kg/yr, respectively. These two species are typically cultivated for timber production and benefit from adequate spacing and better soil nutrient conditions in the study area.

In contrast, *Pinus tabulaeformis* (0.45 kg/yr) and *Ginkgo biloba* (0.07 kg/yr) show limited growth efficiencies. *Pinus tabulaeformis* is a needleleaf species originating from cold climates, which leads to lower utilization of radiative energy compared to broadleaf species. Additionally, both species are occasionally harvested and sold as saplings. Notably, *Populus tomentosa* exhibited a slower-than-expected growth rate, despite being a fast-growing species. This is primarily because a significant portion of *Populus tomentosa* was planted decades ago as part of the shielding trees and now has an older stand age than other species.

## 4. Discussion
### 4.1 Implication of this study
#### 4.1.1 High-Resolution CHM Generation for Plantation Management

Continuously and accurately estimating plantation AGB is a challenging yet essential task for both silviculture and recent carbon abatement activities. This task has gained increased importance as recent initiatives emphasize evaluating the carbon sink potential of afforestation projects, which are typically organized through a series of plantations[8]. Existing methods, such as using VOD at low spatial resolution or optical RS imagery, are not always suitable for this purpose. Current guidelines emphasize the use of airborne lidar observations (China Certified Emission Reduction (CCER), 2022) for quantifying plantation AGB. However, these methods struggle to balance cost and accuracy, as lidar collection requires specialized equipment and intensive human labor. With advancements in computational capacity and computer vision theory, LVFMs pretrained on extensive imagery have proven effective for specific CV tasks. These pretrained networks provide universal features that can be fine-tuned for specific domains, significantly reducing training set requirements and enhancing model performance[33].

This study demonstrates the potential of LFVMs for generating high-resolution CHMs using RGB imagery. The proposed method effectively addresses the limitations of the feature extracted from LVFM using in small plantation parcels, which are often characterized by their compact size and uniform tree height distribution. By introducing a feature enhancement module, the model preserves spatial details that are critical for accurate canopy height and AGB estimation. This capability is particularly beneficial for plantation management, where precise monitoring of tree growth and carbon sequestration is essential for both economic and environmental reasons.

### 4.1.2 Model comparison

Based on the results of Experiment I, our model significantly outperforms existing methods for CHM generation and products (see Section 3.1). A comparison with U-Net demonstrates that our LVFM-based approach surpasses traditional CNN-based methods. The ViT-based network extracts more features by leveraging self-attention mechanisms, enabling it to capture global dependencies across the entire image [52]. Additionally, a comparison with DPT highlights the effectiveness of the self-supervised feature enhancement module, which is the key innovation of this study. This module substantially improves the model's ability to retain fine-grained spatial details, which are crucial for accurate height predictions, addressing the common limitation of naïve LVFMs that tend to lose spatial information. Notably, this enhancement is achieved without the need for additional labels, making it particularly beneficial in plantation settings where high-precision CHM generation is essential. The successful application of this method across different tree species and plantation ages demonstrates its robustness, providing a reliable tool for both current plantation management and future carbon offset projects.

Generalizability constitutes a pivotal criterion in assessing deep learning algorithms, particularly for cross-regional deployment across heterogeneous geographical domains. This study systematically addresses this challenge through rigorous cross-domain validation, i.e., extending the trained network from Fangshan to the ecologically distinct region (Saihanba). The results demonstrate robust model transferability across disparate forest ecosystems, substantiating its generalizability. In contrast, global CHM, such as Tolan's product [42], despite leveraging extensive datasets for planetary-scale mapping, exhibit suboptimal performance in capturing both tree height variability and fine-scale spatial features within our study area. This limitation stems from their inherent optimization for global metrics, which homogenizes localized spectral signatures and spatial patterns, thereby compromising their capacity to resolve ecologically critical sub-meter height and structure variabilities [43]. Our approach, employing fine-tuned LVFM pretrained on natural imagery, demonstrates superior local adaptability. Unlike global models constrained by fixed spectral band configurations, our methodology dynamically prioritizes locally discriminative features, such as crown morphology and height distribution. Furthermore, the hierarchical feature extractors in pretrained models effectively capture multi-scale textural information, outperforming global models optimized for continental-scale features. In conclusion, while LVFMs exhibit robust capabilities in forest structural feature extraction and height estimation, optimal cross-domain application necessitates comprehensive a priori evaluation to mitigate or leverage domain-specific interference.

Additionally, our model also outperforms the mid-resolution CHM products from Lang et al [21]. While these 10-meter resolution CHMs capture global tree height patterns, they perform poorly in reflecting local tree height variability. This limitation arises because NDVI and other canopy greenness indices, derived from optical sensors, fail to bring depth and complex structural information. Our method addresses this shortcoming by utilizing structural features extracted from high-resolution RGB imagery, rather than relying on the relationship between reflectance and structural traits. This comparison underscores the necessity of collecting local data to fine-tune global CHM products for high-standard forestry applications, such as plantation monitoring in this study..

### 4.1.3 AGB estimation and tracking growth

As one of the key applications of tree height estimation, we achieve satisfactory plantation-scale AGB estimation in Experiment II (Section 3.2). According to recent guidelines, a 90% accuracy rate is recommended for tree segmentation (China Certified Emission Reduction (CCER), 2022). Segmenting individual trees from RGB imagery typically requires significant human intervention or the training of a complex computer vision algorithm[53]. However, by using the CHM generated in this study, a local maxima algorithm—without the need for range parameter optimization—meets this criterion for most plantations. This indicates that tree segmentation tasks can be significantly simplified using our CHM results without additional human labour. We then apply the established relationship between tree height and AGB to estimate plantation-scale AGB, achieving satisfactory accuracy for most plantation species. Our estimations closely align with those based on lidar-derived CHM, demonstrating that our method can serve as an effective supplement to lidar. This success is particularly beneficial for future afforestation and management activities, as lidar collection over large regions typically involves substantial investment. The ability to generate comparable CHMs from satellite RGB imagery enables more continuous carbon project monitoring [54]. Additionally, our method offers valuable support for monitoring projects that rely on optical remote sensing imagery, yet it is based on a fundamentally different theory from traditional fitting-based methods [55]. Instead of directly utilizing optical reflectance, we extract structural and depth features to estimate AGB, thereby avoiding potential bias caused by saturation effects in dense canopy regions [14].

Tracking growth and estimating annual AGB for regional forests has historically been a challenging task, particularly for plantations with significant economic or environmental benefits, which is the focus of our Experiment II [56]. Current studies relying on optical remote sensing imagery or VOD often fall short in fully capturing carbon storage traits, leading to controversial conclusions about terrestrial carbon sink capacity. For instance, assuming a constant relationship between canopy water content and AGB introduces unacceptable variability in annual AGB estimation [57]. Our method offers a novel and promising approach to addressing these issues by providing an AGB benchmark based on accurate CHM generation. As demonstrated by our results, we achieve CHM with an approximate 0.2-meter standard error in estimating tree heights for plantations, indicating the method's effectiveness in detecting annual growth, which is typically less than 1 meter during the early stages of development. The reasonable annual AGB estimations presented in Section 3.3 further illustrate this potential. Moreover, our method requires only a minimal training set and high-resolution RGB imagery from satellites during the inference stage, making data collection more feasible. This advantage reduces the reliance on extensive lidar samples, enabling the generation of annual CHMs at regional or national scales with minimal data. This capability is crucial for evaluating carbon abatement achievements and enhancing our understanding of regional carbon cycling.

### 4.2 Limitations and Future Research Directions

Despite the success of this study in estimating tree heights and AGB, several challenges remain before broader application can be realized. First, a more comprehensive dataset is needed for model evaluation. This study was conducted in the region characterized by relatively homogeneous plantations; however, the existing LVFM-based method was designed primarily for global tree height mapping. It remains unclear whether our newly designed feature enhancement module will perform adequately in natural forests, where canopy structures are more irregular and crown architectures are less uniformed. To address this, lidar observations should extend beyond plantations to include natural forests with varying canopy coverages and topographical conditions. Furthermore, additional testing is required to determine whether our method can be further improved to mitigate the "pollution" from globally trained models.

Second, a more rigorous evaluation is necessary before large-scale detection of plantation or forest growth can be considered, which forms the basis of improving large scale carbon budget. In this study, our method for tracking tree height and AGB is based on a qualitative relationship with stand age, rather than on rigorous training and validation using field observations. As a result, the accuracy and reliability of growth tracking in this study are somewhat limited. This limitation arises from the challenges of obtaining annual lidar measurements or historical field data. Collecting multi-source data, such as records from local forestry bureaus or forest inventories [58], would significantly improve the robustness of growth tracking and increase confidence in its effectiveness.

Finally, two critical steps in AGB calculation—single-tree segmentation and species identification—were simplified in this study, involving substantial manual intervention. As Brandt et al. [59] has suggested, typical CNN-based networks are effective for single-tree segmentation, and the ViT-based method further enhances model capabilities [60]. However, accurate segmentation of tree crowns in closed canopy environments, such as the *R. Pseudoacacia* plantations in current study area, is fundamental to precise AGB estimation. Additionally, while tree species classification methods have proven effective for supporting large-scale forest investigation [61, 62], it remains uncertain whether these methods can meet the specific requirements of plantation monitoring and to what extend LVFM-extracted features can inform classifiers warrants further evaluation.

## 5. Conclusion

As plantations are considered an adequate solution for mitigating climate risks, estimating their AGB is a prerequisite for evaluating their effectiveness. Typical RS-based methods are not suitable for estimating AGB in plantations with small areas and fuzzy boundaries. Traditional DL methods for CHM generation, the results of which are employed in AGB estimation, usually require unacceptably large training datasets. In this study, we present a novel LVFM-based approach for CHM generation that offers a promising alternative to traditional methods, balancing cost, accuracy, and scalability. By enhancing the spatial resolution of extracted features and reducing reliance on extensive training datasets, this method provides a practical tool for plantation monitoring and carbon sequestration assessment. Two experiments were conducted to evaluate the model's performance and its applicability in monitoring plantations in a typical plantation-establishing region of northern China. Based on pixel-wise evaluation and visual inspection, the results of Experiment I demonstrate that DINOv2-Large, serving as the core feature extractor in our model, generates the best feature maps for downstream tasks, exhibits satisfactory generalizability, and outperforms other DL models and existing CHM products. Experiment II shows that the resulting CHM effectively supports single-tree segmentation and AGB estimation for plantations, as our method preserves considerable fine-grained features from RGB imagery. Additionally, the generated CHM can be used to track annual growth in plantations with diverse species. Our study presents a promising approach for generating high-resolution CHMs from RGB imagery and evaluating their potential application in plantation and forestry monitoring, particularly for assessing CCER afforestation projects in the near future. Future work should focus on expanding its applicability to diverse forest types and integrating automated classification techniques to fully realize its potential for large-scale forestry applications.

**Acknowledgments**

**General**: Thank others for any contributions.

**Author contributions:** S.T.: Data acquisition, formal analysis, and writing—original draft. X.Zhang.: Methodology, formal analysis, and writing. L.X.H.: Methodology, writing and reviewing. H.G.H., H.W.: writing and reviewing.

**Funding:** This study is supported by the National Science Foundation of China (72140005), Natural Science Foundation of Beijing, China (Grant No. 3252016), and partly by BBSRC(BB/R019983/1, BB/S020969/), and EPSRC(EP/X013707/1), Key Research and Development Program of Shaanxi Province (Program No. 2024NC-YBXM-220).

**Competing interests:** The authors declare that there is no conflict of interest regarding the publication of this article.

**Data Availability:** The data and code that support the findings of this study are available from the corresponding author upon reasonable request.